\documentclass[onecolumn]{fairmeta}
\usepackage[utf8]{inputenc}
\usepackage[dvipsnames]{xcolor}
\usepackage{pifont}
\usepackage{xcolor}  
\usepackage{hyperref,xurl}
\usepackage{graphicx}
\usepackage{booktabs}
\usepackage{makecell}
\usepackage{caption}
\usepackage{wrapfig}
\usepackage{float}
\usepackage{amssymb}

\newcommand{\xmark}{\ding{55}}

\usepackage{hyperref}
\usepackage{url}
\urldef{\urlrbm}\url{https://www.meta.com/ai-glasses} 
\urldef{\urlrabbit}\url{https://www.rabbit.tech/rabbit-r1}
\urldef{\urlhumane}\url{https://humane.com/}
\urldef{\urlcragmm}\url{https://huggingface.co/crag-mm-2025}

\usepackage[most]{tcolorbox}
\newtcolorbox[list inside=prompt,auto counter]{prompt}[1][]{
    breakable,
    enhanced jigsaw,
    colbacktitle=black!60,
    coltitle=white,
    fontupper=\footnotesize,
    boxsep=5pt,
    left=0pt,
    right=0pt,
    top=0pt,
    bottom=0pt,
    boxrule=1pt,
    #1,
}
\usepackage{listings}
\title{CRAG-MM: Multi-modal Multi-turn Comprehensive RAG Benchmark}

\author[1, \dagger]{Jiaqi Wang}
\author[1, \dagger]{Xiao Yang}
\author[1]{Kai Sun}
\author[1]{Parth Suresh}
\author[1]{Sanat Sharma}
\author[1, *]{Adam Czyzewski}
\author[1]{Derek Andersen}
\author[1]{Surya Appini}
\author[1]{Arkav Banerjee}
\author[1]{Sajal Choudhary}
\author[1]{Shervin Ghasemlou}
\author[1, *]{Ziqiang Guan}
\author[1]{Akil Iyer}
\author[1]{Haidar Khan}
\author[1]{Lingkun Kong}
\author[1, *]{Roy Luo}
\author[1]{Tiffany Ma}
\author[2]{Zhen Qiao}
\author[1]{David Tran}
\author[1]{Wenfang Xu}
\author[1]{Skyler Yeatman}
\author[1]{Chen Zhou}
\author[1]{Gunveer Gujral}
\author[4]{Yinglong Xia}
\author[1]{Shane Moon}
\author[1]{Nicolas Scheffer}
\author[1]{Nirav Shah}
\author[1]{Eun Chang}
\author[1]{Yue Liu}
\author[1]{Florian Metze}
\author[1]{Tammy Stark}
\author[1]{Zhaleh Feizollahi}
\author[1]{Andrea Jessee}
\author[1]{Mangesh Pujari}
\author[1]{Ahmed Aly}
\author[1]{Babak Damavandi}
\author[1]{Rakesh Wanga}
\author[1]{Anuj Kumar}
\author[2]{Rohit Patel}
\author[3]{Wen-tau Yih}
\author[1]{Xin Luna Dong}
\affiliation[1]{Meta Reality Labs}
\affiliation[2]{Meta Superintelligence Labs}
\affiliation[3]{FAIR, Meta}
\affiliation[4]{Meta}

\contribution[\dagger]{Equal contribution}
\contribution[*]{Work done at Meta}

\abstract{Wearable devices such as smart glasses are transforming the way people interact with their surroundings, enabling users to seek information regarding entities in their view. {\em Multi-Modal Retrieval-Augmented Generation (MM-RAG)} plays a key role in supporting such questions, yet there is still no comprehensive benchmark for this task, especially regarding wearables scenarios. To fill this gap, we present \textbf{CRAG-MM}---a Comprehensive RAG benchmark for Multi-modal Multi-turn conversations. CRAG-MM contains a diverse set of 6.5K (image, question, answer) triplets and 2K visual-based multi-turn conversations across 13 domains, including 6.2K {\em egocentric} images designed to mimic captures from wearable devices. We carefully constructed the questions to reflect real-world scenarios and challenges, including five types of image-quality issues, six question types, varying entity popularity, differing information dynamism, and different conversation turns. We design three tasks: single-source augmentation, multi-source augmentation, and multi-turn conversations---each paired with an associated retrieval corpus and APIs for both image-KG retrieval and webpage retrieval. Our evaluation shows that straightforward RAG approaches achieve only 32\% and 43\% {\em truthfulness} on CRAG-MM single- and multi-turn QA, respectively, whereas state-of-the-art industry solutions have similar quality (32\%/45\%), underscoring ample room for improvement. The benchmark has hosted KDD Cup 2025, attracting about 1K participants and 5K submissions, with winning solutions improving baseline performance by 28\%, highlighting its early impact on advancing the field.

}

\metadata[Benchmark]{\url{https://huggingface.co/crag-mm-2025}}
\correspondence{First Authors at \email{jqwang@meta.com}, \email{xiaoyangfb@meta.com}}

\begin{document}

\maketitle

\section{Introduction}
{\em Wearable AI devices} are revolutionizing how people interact with computing systems. Modern wearable devices, such as Rayban Meta~\footnote{\urlrbm}, Rabbit R1~\footnote{\urlrabbit}, and the Humane AI Pin~\footnote{\urlhumane}, enable vision-based conversations where users can ask questions about objects in their view. For example, a user may inquire about the history of a landmark they are viewing, the price of a product they are holding on different e-commerce websites, or repair instructions for a broken household device. What such questions often have in common is {\em the need for factual information that cannot be inferred from the images alone}, necessitating multi-modal Retrieval-Augmented Generation (MM-RAG) systems that access external sources for enriched and accurate responses. To advance the MM-RAG techniques, we construct the {\sc CRAG-MM} benchmark, a comprehensive multi-modal benchmark of 6.5K single-turn and 2K multi-turn conversations, with an emphasis on use cases relevant to wearable AI devices. Specifically, our motivation is three-fold.


First, although many Visual Question Answering (VQA) benchmarks exist~\citep{antol2015vqa, hudson2019gqa, schwenk2022okvqa}, they rely primarily on common knowledge~\citep{antol2015vqa, goyal2017making, schwenk2022okvqa} and visual reasoning~\citep{hudson2019gqa}, which is insufficient to assess factual questions comprehensively. New benchmarks focusing on multi-modal knowledge integration have emerged in recent years. They nevertheless were constructed solely from Wikipedia~\citep{lerner2022viquae}, created by templates~\citep{talmor2021multimodalqa}, or consisted of questions uncommon in real life~\citep{talmor2021multimodalqa, chang2022webqa}. {\sc CRAG-MM} addresses these limitations by incorporating 8K conversations, with ~89\% focusing on factual questions that require external information for trustworthy answers. 

Second, several question-answering (QA) benchmarks have emerged in recent years, covering {\em closed-book} (e.g., SimpleQA~\citep{wei2024measuring}, FACTS Grounding~\citep{jacovi2025facts}) and {\em open-book} (e.g., FreshQA~\citep{vu2023freshllms}) QA. Some, like CRAG~\citep{yang2024crag}, specifically target the evaluation of RAG systems. {\sc CRAG-MM} goes beyond them not only by introducing {\em multi-modal} use cases, but also by including 2k {\em multi-turn conversations}, where ${\sim}38\%$ of them involve domain shifts. This simulates natural topic drift, a common feature of human conversations, further enhancing the realism of the benchmark. 

Third, existing VQA benchmarks typically feature high-quality images. In contrast, wearable devices often use wide-angle cameras, capturing {\em egocentric} images where objects of interest appear small, rotated, truncated, occluded, blurred, or poorly lit~\citep{shenoy2024lumos}. To bridge this gap, the {\sc CRAG-MM} benchmark includes 7.9K images, with 79\% being egocentric, reflecting the real-world challenges in wearable AI applications.

As such, {\sc CRAG-MM} is, to the best of our knowledge, {\em the first publicly released benchmark capable of effectively evaluating multi-modal RAG} and also {\em one of the earliest benchmarks designed to reflect wearable AI use cases}. The design of {\sc CRAG-MM} follows the key features outlined in~\citep{yang2024crag}, with the following principles.

\begin{figure}[t]
  \centering
  \includegraphics[width=0.9\columnwidth]{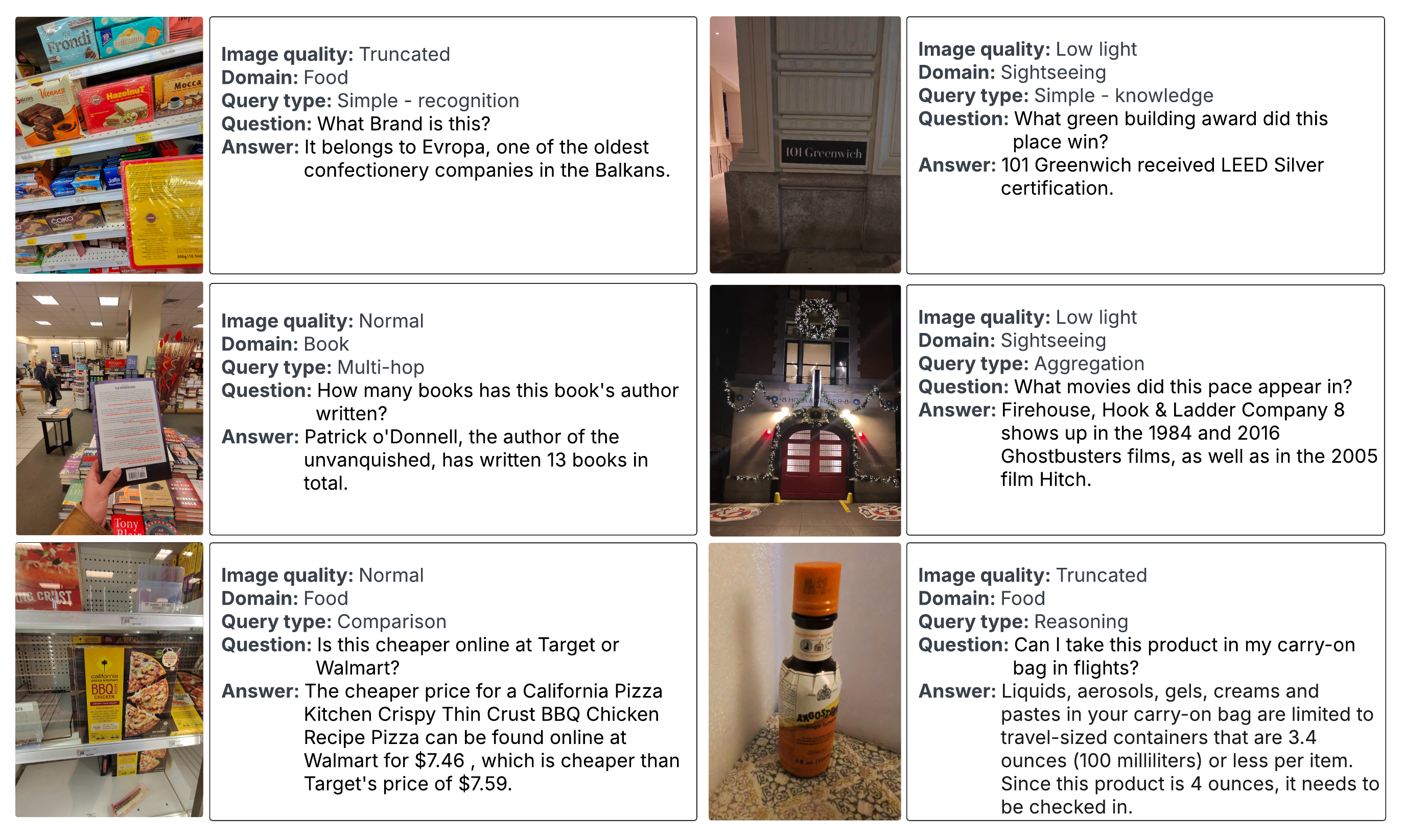}
  \captionsetup{justification=centering}
  \caption{CRAG-MM examples.}
  \label{fig:crag_mm_example_1}
\end{figure}

\smallskip
\noindent
{\bf Rich and insightful benchmark:} {\sc CRAG-MM} is structured along four key dimensions---image quality, question type, entity popularity, and conversation complexity, mirroring real-world challenges faced by wearables QA systems and enabling in-depth analysis and debugging. It covers both common and challenging use cases; for example, 15\% of images are low-quality egocentric images, 21\% questions involve torso-to-tail entities; 52\% questions are complex questions and require multi-source information synthesis, and 23\% conversations contain multiple turns. 
See Fig.\ref{fig:crag_mm_example_1} for some examples from {\sc CRAG-MM}.

\smallskip
\noindent
{\bf Reliable and fair evaluation:} To ensure fair comparison, we provide equal access to retrieval resources, including APIs for an image knowledge graph (KG) with 68K entries and a webpage corpus with 800K pages. The retrieval corpus is designed to mimic real-world conditions,
with a 1:20 ratio of relevant to irrelevant information for image search and 1:2 for web search, 
allowing comparison of different solutions fairly and easily. 
The retrieval corpus, search APIs, and relevant metadata allow for evaluating different components of RAG systems: entity recognition, OCR, query rewrite, response generation, and so on.



\smallskip
\noindent
{\bf Challenging questions:} CRAG-MM contains realistic but hard questions. As shown in Tab.\ref{tab:baseline_benchmarking_full}, without RAG, the most advanced multi-modal large language model (MM-LLM) we evaluated (GPT-5 mini) achieved an accuracy of 37\% on single-turn questions; straightforward solutions improved the accuracy to 50\%; even the most advanced industry solutions (GPT-5), with access to potentially richer web corpus and better search engine, reached only an accuracy of 63\% with 31\% hallucinations. Multi-turn conversations contain simpler questions; still, the best industry solution achieved an accuracy of 70\%, with 27\% conversations being early stopped due to two consecutive erroneous or missing answers. 

The {\sc CRAG-MM} benchmark served as the foundation for the KDD Cup 2025 Challenge\footnote{\url{https://www.aicrowd.com/challenges/meta-crag-mm-challenge-2025}}, attracting nearly 1K participants and 5K submissions. The winning solution improved over straightforward solutions by 28\%, highlighting its early impact on advancing the field. 

\smallskip
\noindent
{\bf Differences from CRAG and other existing benchmarks:} 
To the best of our knowledge, CRAG-MM is the first comprehensive benchmark that focuses on wearable AI use cases. 
Different from CRAG~\citep{yang2024crag} which targets text-based single-turn QA, CRAG-MM is a visual QA benchmark. 
Unlike other existing VQA benchmarks, CRAG-MM uniquely features wearable use cases -- questions are based on egocentric images and often require external knowledge to answer. It encompasses a variety of domains and question types, which can help reveal interesting insights and facilitate development.  
Further, CRAG-MM extends beyond single-turn QA by including multi-turn conversations, a common and essential use case for wearable devices. Tab.\ref{tab:benchmark_comparison} compares CRAG-MM to a few popular or recent benchmarks.

\begin{minipage}{\textwidth}
\centering
  \small
  \label{tab:benchmark_comparison}
  \captionof{table}{Comparing CRAG-MM to existing benchmarks for factual question answering: SnapNTell~\citep{qiu2024snapntell}, WebQA~\citep{chang2022webqa}, MultiModalQA~\citep{talmor2021multimodalqa}, MM-Vet~\citep{yu2023mm}, MT-Bench-101~\citep{bai2024mt}, CRAG~\citep{yang2024crag}.}
  \begin{tabular}{lcccccc}
    \toprule
    \textbf{Benchmark} & \textbf{\makecell{Ego-\\centric}} & \textbf{\makecell{Image search\\API}} & \textbf{\makecell{Web search\\API}} & \textbf{\makecell{Multi-\\modal}} & \textbf{\makecell{Multi-\\turn}} & \textbf{\makecell{Torso \&\\tail facts}} \\
    \midrule
    \makecell[l]{SnapNTell} & \color{red}\xmark & \color{red}\xmark & \color{red}\xmark & \color{Green}\checkmark & \color{red}\xmark & \color{Green}\checkmark  \\
    
    \makecell[l]{WebQA} & \color{red}\xmark & \color{Green}\checkmark \footnote{Prefetched images.} & \color{Green}\checkmark \footnote{Prefetched snippets from wikipedia.} & \color{Green}\checkmark & \color{red}\xmark & \color{red}\xmark  \\
    
    \makecell[l]{MultiModalQA} & \color{red}\xmark & \color{red}\xmark & \color{red}\xmark & \color{Green}\checkmark & \color{red}\xmark & \color{red}\xmark  \\
    
    \makecell[l]{MM-Vet} & \color{red}\xmark & \color{red}\xmark & \color{red}\xmark & \color{Green}\checkmark & \color{red}\xmark & \color{red}\xmark  \\
    
    \makecell[l]{MT-Bench-101} & \color{red}\xmark & \color{red}\xmark & \color{red}\xmark & \color{red}\xmark & \color{Green}\checkmark & \color{red}\xmark  \\

    \makecell[l]{CRAG} & \color{red}\xmark & \color{red}\xmark & \color{Green}\checkmark & \color{red}\xmark & \color{red}\xmark & \color{Green}\checkmark \\
    
    \makecell[l]{CRAG-MM} & \color{Green}\checkmark & \color{Green}\checkmark & \color{Green}\checkmark & \color{Green}\checkmark & \color{Green}\checkmark & \color{Green}\checkmark \\
    \bottomrule
  \end{tabular}
\end{minipage}

\section{Problem description}

An {\em MM-RAG QA system} takes as input an image $I$ and a question $Q$, and outputs an answer $A$; the answer is generated by MM-LLMs based on information retrieved from external sources, combined with knowledge internalized in the model. 
A {\em Multi-turn MM-RAG QA system} in addition takes questions and answers from previous turns as context to answer new questions. The answer should provide useful information to answer the question, without adding any hallucination. 

\subsection{Question types}
We first define six types of questions  as follows. 

\begin{itemize}
    \item {\em Simple-recognition}: Questions asking for simple facts that can be directly answered from the image, e.g., {\em ``what brand is the milk''} or {\em ``who wrote this book''}, where the brand name and the book author are shown on the image.
    \item {\em Simple-knowledge}: Questions asking for simple facts that require external knowledge to answer, e.g., {\em ``what's the price of this sofa on Amazon''}. 
    \item {\em Multi-hop questions}: Questions that require chaining multiple pieces of information to compose an answer, such as {\em ``what other movies has the director of this movie directed''}.
    \item {\em Comparison questions}: Questions require comparing multiple pieces of information, such as {\em ``is this cheaper on Amazon''} (where the image is showing a product and its store price). 
    \item {\em Aggregation questions}: Questions require aggregating multiple pieces of information, {\em ``which drinks do not contain added sugar among these''} (where the image is showing a few drinks in a grocery store).
    \item {\em Reasoning questions}: Questions about an entity that cannot be directly looked up from the retrieved content and require reasoning to answer, such as {\em ``can the dryer be used in Europe''} (where the image shows a dryer).
\end{itemize}

\begin{figure}[t]
  \centering
  \includegraphics[width=\columnwidth]{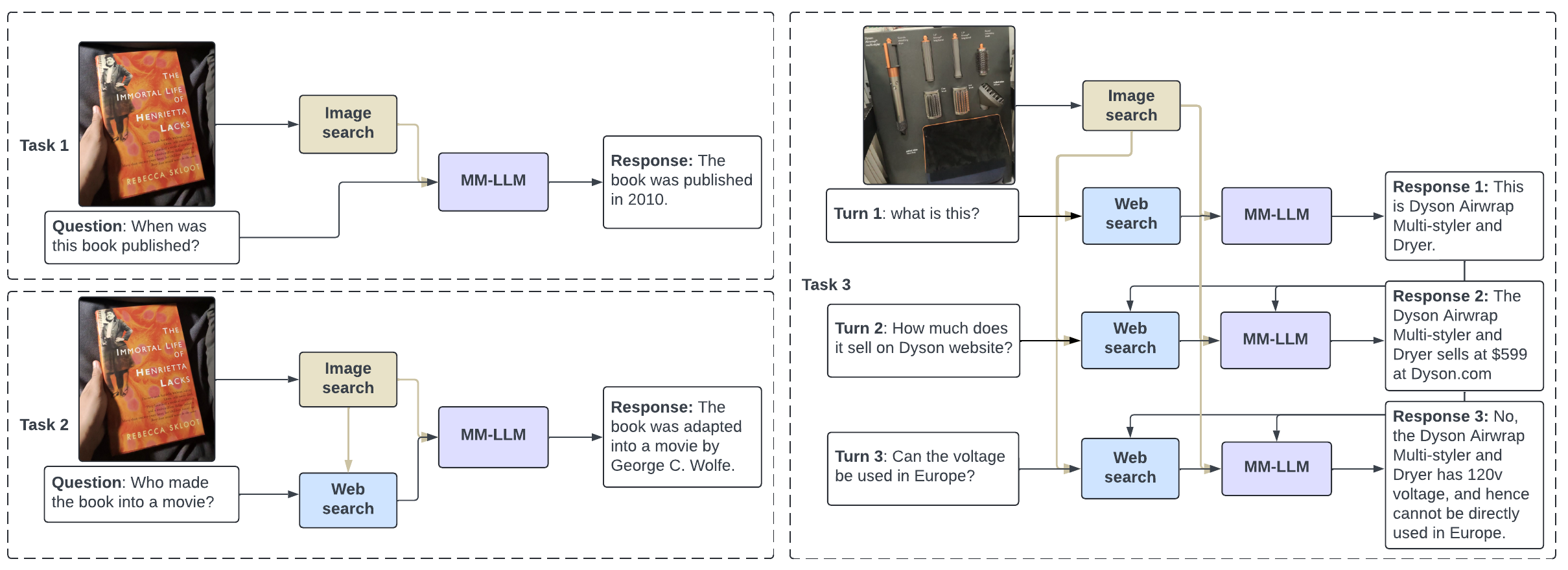}
  \captionsetup{justification=centering}
  \caption{CRAG-MM task design.}
  \label{fig:competition_task}
\end{figure}

\subsection{Tasks}
We designed three tasks. As shown in Fig.\ref{fig:competition_task}, Task 1 and Task 2 contain single-turn questions, where the former provides image-KG-based retrieval, and the latter additionally introduces web retrieval; Task 3 focuses on multi-turn conversations. Here, we provide the content that can be leveraged in QA to ensure a fair competition. We describe how we generated the data in the next section. The three tasks can guide users of the benchmark in building end-to-end MM-RAG systems.

\smallskip
\noindent
\textbf{Task 1: Single-source augmentation.} Task 1 provides an image search {\em mock API} to access information from an underlying image-based {\em mock KG}. The mock KG is indexed by the image, and stores structured data associated with the image; answers to the questions may or may not exist in the mock KG. The mock API takes an image as input, and returns similar images from the mock KG along with metadata associated with each image to support answer generation. This task aims to test basic answer generation capability of MM-RAG systems.

\smallskip
\noindent
\textbf{Task 2: Multi-source augmentation.} In addition to image search, Task 2 provides a web search mock API as retrieval source. The webpages likely contain useful information for answering the question, but also contain noise. This task aims to test how well the MM-RAG system synthesizes information from different sources. 

\smallskip
\noindent
\textbf{Task 3: Multi-turn QA.} Task 3 tests the system's ability to conduct multi-turn conversations. Each conversation contains 2--6 turns. Questions in second turn or beyond may or may not need the image for answering the questions. Task 3 tests context understanding for smooth multi-turn conversations.

\section{Data description}
\label{sec:dataset_description}
We now describe the benchmark dataset. CRAG-MM covers 13 domains and six types of questions, all in English. It contains three parts of data: the image set, the QA set, and the contents for retrieval. 

\subsection{Image set}
CRAG-MM contains two types of images: {\em egocentric} and {\em normal}. The egocentric images were collected using smart glass 
devices from first-person perspective; the normal images were collected from publicly available images. We collected both types of images via two methods: {\em pre-defined collection} and {\em free-form collection}. 

For pre-defined collection, we first specified a list of domains and entity types likely to be involved in user wearable interactions, such as {\em electronic products} in the {\em shopping} domain. We adopted the method described in~\citet{sun2023head} and sampled entities of top, middle and bottom popularity.
We defined the popularity based on some publicly available traffic information for each entity type, and created equal number of questions for each bucket. We recruited vendors to collect images for the sampled entities using the RayBan Meta Smart Glasses\footnote{\urlrbm} whenever possible; otherwise, we searched online to find an image for the target entity from public sources. 

For free-form collection, we asked the vendors to wear a pair of smart glasses and interact in their daily lives. We recommended to collect $80\%$ images from the provided domains, and $20\%$ for any additional use cases that the vendors found plausible for interaction with smart glasses. 

We collected 6.2K egocentric images and 1.7K public images in total. We also requested $15\%$ of the egocentric images to be captured under imperfect conditions: low-light, blurred, truncation, occlusion, or rotation. See  Tab.\ref{tab:image_quality_definition} in Appendix~\ref{appendix:image_quality_definition} for the definition and Tab.\ref{tab:image_statistics} for the distribution.


\begin{table}[t]
    \centering
    \captionsetup{type=table, justification=centering, singlelinecheck=false}
    \captionof{table}{Distribution by image quality.}
    \small
    \begin{tabular}{l|rr|r}
        \toprule
        \textbf{Type} & \textbf{Egocentric} & \textbf{Normal} & \textbf{Total} \\
        \midrule
        Normal & 5143 & 1593 & 6736     \\
        Low-light & 267 & 33 & 300    \\
        Blurred & 191 & 24 & 215   \\
        Truncated & 370 & 31 & 401     \\
        Occluded & 116 & 8 & 124   \\
        Rotated & 161 & 6 & 167  \\
        \midrule
        Total & 6248 & 1695 & 7943 \\
        \bottomrule
    \end{tabular}
    \label{tab:image_statistics}
\end{table}

\subsection{Question Answering data} 
\textbf{Single-turn QA data:} We constructed the QA set for the collected images based on two sources---KGs or web contents. For the QA set constructed from KGs, we first leveraged existing entity types and relations in the KGs to create meaningful QA templates. We then used entities sampled from the image collection to pair with the question templates and took the associated attribute values as the answer. For the QA set constructed from web contents, we asked annotators to create plausible questions for wearable devices that could possibly be answered by web search. The annotators then also created the complete web search query based on the image and the question, and recorded the ground truth answers. See Appendix~\ref{appendix:qa_data_creation} for more details.   


\textbf{Multi-turn QA data:} We first created a number of simple questions based on the web contents as first-turn seed questions. We then prompted \texttt{Llama-3.2-90B-Vision-Instruct}~\citep{dubey2024llama, meta2024llama32} to create multi-turn conversations based on the given seed questions. Each conversation has two to six turns, and covers one to three domains. Next, we asked the annotators to review the conversation, remove or revise turns that are not plausible, too simple, or do not have a single indisputable answer. Finally, the annotators also created the ground truth answers for each turn by conducting web search.

\begin{table*}[t]
  \small
  \begin{center}
  \caption{Distribution of domains and question types for CRAG-MM single- and multi-turn QA. We use the first turn's domain for multi-turn conversation.}
  \label{tab:data_statistics_type}
      \begin{tabular}{l|rrrrrr|r|r}
        \toprule
        & \textbf{\makecell{Simple-\\rec.}} & \textbf{\makecell{Simple-\\know.}} & \textbf{\makecell{Multi-\\hop}} & \textbf{\makecell{Comp.}} & \textbf{\makecell{Agg.}} & \textbf{\makecell{Reason.}} & \textbf{\makecell{Single-\\turn}} & \textbf{\makecell{Multi-\\turn}}\\
        \midrule
        Animal & 10 & 191 & 67 & 115 & 71 & 66 & 520 & 148  \\
        Book & 12 & 53 & 37 & 17 & 12 & 9 & 140 & 147 \\
        Food & 35 & 312 & 108 & 132 & 99 & 85 & 771 & 196 \\
        General Obj. Rec. & 26 & 210 & 82 & 74 & 43 & 50 & 485 & 184 \\
        Local & 29 & 428 & 168 & 119 & 106 & 87 & 937 & 181 \\
        Math \& Science & 61 & 62 & 16 & 17 & 18 & 20 & 194 & 130 \\
        Plants \& Gardening & 19 & 443 & 119 & 125 & 117 & 107 & 930 & 118 \\
        Shopping & 52 & 239 & 48 & 66 & 41 & 48 & 494 & 204 \\
        Sports \& Games & 12 & 131 & 31 & 24 & 21 & 12 & 231 & 123 \\
        Style \& Fashion & 13 & 57 & 15 & 14 & 5 & 9 & 113 & 115 \\
        Text Understanding & 170 & 113 & 15 & 10 & 21 & 18 & 347 & 113 \\
        Vehicle & 14 & 250 & 147 & 196 & 113 & 140 & 860 & 120 \\
        Other & 9 & 125 & 75 & 68 & 105 & 58 & 440 & 177 \\
        \midrule
        Total & 462 & 2614 & 928 & 977 & 772 & 709 & 6462 & 1956 \\
        \bottomrule
      \end{tabular}
  \end{center}
\end{table*}

We collected 6.5K single-turn questions and 2K multi-turn conversations with an average length of 4.9 turns in the final dataset. Tab.\ref{tab:data_statistics_type} summarizes distribution of the questions across different dimensions. The size of each slice allows us to get metrics with ${<}5\%$ margin-of-error (with 95\% confidence level) for most of the slices.

\subsection{Retrieval contents}
\label{sec:retrieval_contents}
We included two types of content for retrieval to simulate the practical scenarios for MM-RAG: image-based KG search and text-based web search. We carefully designed the corpus to highlight the real challenges a retrieval system faces and included potential noises in the retrieval results.

\textbf{Image-based KG search:} Image search is a key component of MM-RAG systems. To simplify, we limit image search to KG-search. We provide an API that takes an image as input and returns similar images with associated structured metadata from a mock KG using \texttt{CLIP ViT-L/14@336px}~\citep{radford2021clip}. 
Our image KG covers 93\% of the entities referenced in the questions, and querying the index with original images achieves only 52\% recall (Fig.\ref{fig:image_and_web_search_recall}). See Fig.\ref{fig:image_mock_api_example} in  Appendix~\ref{appendix:dataset} for an example and Appendix~\ref{appendix:image_search_api_creation} for details on how we created the API.

\textbf{Text-based web search: }
\label{sec:web_search_result}
We also construct web search to obtain additional information for question answering. We provide a text-search API that takes a text query as input and returns relevant webpages, with an estimated recall of 89\% (Fig.~\ref{fig:image_and_web_search_recall}). We chunk the documents into 512-token chunks and embed them using BGE~\citep{xiao2024c}. At inference, queries are matched against the chunks to return corresponding webpages.
See Appendix~\ref{appendix:web_search_api_creation} for details about the web search API creation.

\begin{figure}[h]
    \centering
    \includegraphics[width=\linewidth, keepaspectratio]{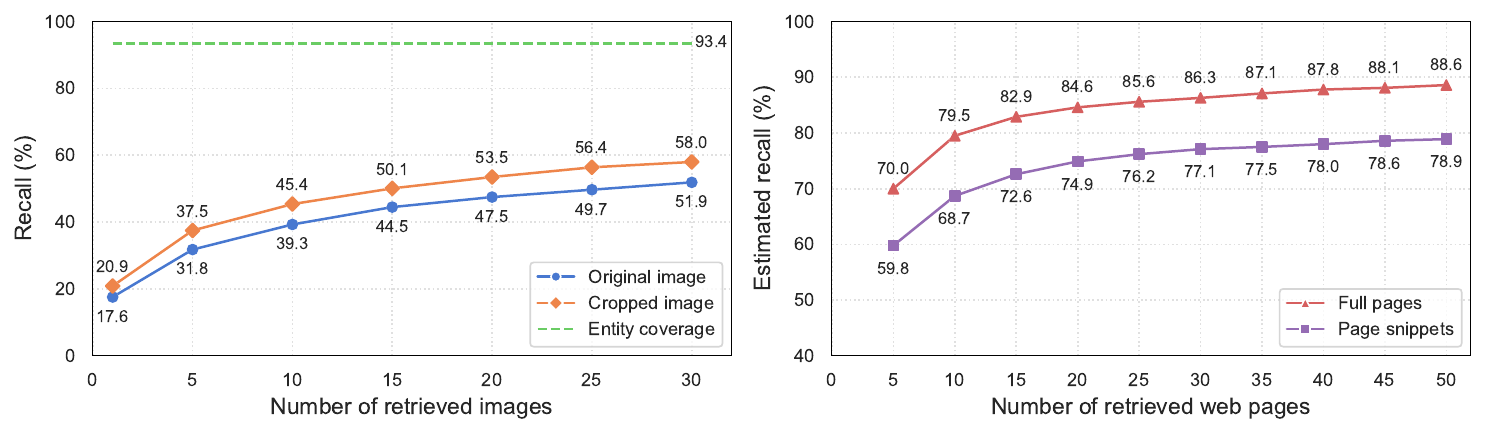}
    \captionsetup{width=\linewidth}
    \captionof{figure}{\textbf{Image search recall (left)}: the image search index covers 93.4\% of the query entities. However, using the full image to query the index only achieves 51.9\% recall. Manual cropping improves the recall slightly to 58.0\%. 
    \textbf{Web search recall (right)}: the top 50 web search results are estimated  to contain the ground truth facts for 88.6\% questions.}
    \label{fig:image_and_web_search_recall}
\end{figure}

\subsection{Data split}
\label{sec:evaluation_test_sets}
The above methods gave us 800K webpage urls (with 2.7M chunks) and a KG of 68K images and 26K entities. We split the data in two steps. We first held out two domains for private test only, for testing the generalization of the MM-RAG systems. We then split the remaining domains' data into three sets with similar distributions: {\em validation, public test}, and {\em private test}, containing 30\%, 30\%, 40\% of the data respectively. We released the validation and public test sets on HuggingFace\footnote{\url{https://huggingface.co/crag-mm-2025}}.

\section{Metrics and Evaluation}
\label{sec:evaluation}
We adopt the metrics and evaluation in 
~\citet{yang2024crag} and extend them for multi-turn QA.


\textbf{Single-turn QA:} We score the model response with \textit{perfect} (fully correct), \textit{missing} (e.g., “I don’t know”, “I'm sorry I can't find ...”), or \textit{incorrect} (wrong or irrelevant) answers as $1$, $0$, and $-1$, and compute \textbf{Truthfulness} as the average score. 

\textbf{Multi-turn QA:}
There is no dominant way to evaluate multi-turn conversations. We adapt the method in~\citet{bai2024mt}, which is mostly closely aligned with information-seeking tasks (as opposed to task fulfilling). In particular, we \textit{early stop} a conversation when the answers for two consecutive turns are wrong or missing, and consider the remaining turns in the same conversation as missing---mimicking the behavior of users after seeing repeated failures. We then calculate the average score of all turns as the \textbf{Truthfulness} for the multi-turn session, and take the grand average for all conversations.

\textbf{Auto-evaluation:}
We adopt LLM-as-a-judge to evaluate the quality of the answers, 
which achieves an accuracy of 99.1\% (Tab.\ref{tab:auto-eval_accuracy} in Appendix~\ref{appendix:evaluation}).


\section{Benchmarking}
\label{sec:benchmarking}
In this section, we present the performance of visual QA systems on the CRAG-MM public test set, demonstrating that CRAG-MM can provide insights and indicate directions for developing MM-RAG techniques. We answer the following three research questions (RQ) through our experiments.


{\bf RQ1:} Does {\sc CRAG-MM} reveal new challenges beyond straightforward solutions?

{\bf RQ2:} How well do state-of-the-art industry solutions perform on {\sc CRAG-MM}?

{\bf RQ3:} What directions for quality improvement are suggested by the {\sc CRAG-MM} benchmark?

\subsection{RQ1: New Challenges Beyond Straightforward MM-RAG Solutions}
\label{sec:baseline_benchmarking}
We present results for three models in this section and leave an extensive comparison in Appendix~\ref{appendix:baseline_eval_full}. The three models are: \texttt{Llama-3.2-90B-Vision-Instruct}, 
\texttt{GPT-5-mini} \citep{openai2025gpt5} and \texttt{Gemini-2.5-Flash} \citep{comanici2025gemini}. 
We implemented an MM-LLM-only solution and three straightforward RAG solutions.
See  Appendix~\ref{appedix:baseline_mmrag_implementation} for more details.

\begin{itemize}
    \item {\bf MM-LLM-only:} We employ a simple prompting strategy that encourages concise answers and the use of {\em ``I don't know''} when the confidence is low. 
    \item {\bf Task 1:} Our Task 1 solution uses the entire image to query the image search API and retrieves top 30 results. After applying a threshold of 0.75 to the similarity score, the retrieved metadata are concatenated to fill in a fixed size context window of 2K tokens.
    \item {\bf Task 2:} Our Task 2 solution incorporates query rewriting for web search and concatenates webpage snippets as additional reference text, expanding the context window up to 8K tokens (similar to~\citet{kandpal2023large} and \citet{mallen2023}).
    \item {\bf Task 3:} In addition to the multi-source retrieval content, our Task 3 solution includes full conversation history to provide context for the current turn.
\end{itemize}


\begin{table*}[t]
\centering
\caption{Performance of straightforward  solutions on CRAG-MM single- and multi-turn QA. All numbers are in percentage. \textbf{Bold} for best quality and \textit{Italic} for best quality within the group. Even the best MM-LLM achieves only 18\% truthfulness for single-turn QA and 30\% for multi-turn QA; straightforward MM-RAG solutions improve only to 32\% for single-turn and 43\% for multi-turn.}
\small
\begin{tabular}{lllrrrrrr}
\toprule
  & & \textbf{Model} & \textbf{Acc.} & \textbf{Miss.} & \textbf{Hallu.} & \textbf{Truth.} & \textbf{Early Stop.}   \\
\midrule
\textbf{Single-turn} & \textbf{MM-LLM}   
                  & Llama 3.2 90B  & 28.2 & 39.1 & 32.8 & -4.6 & - \\
                  & & Gemini 2.5 Flash  & 36.6 & \textit{38.0} & 25.4 & 11.2 & -  \\  
                  & & GPT-5 Mini   & \textit{37.4} & 43.7 & \textit{19.0} & \textit{18.4} & -  \\
\cmidrule{2-8}
& \textbf{Task 1}   
                  & Llama 3.2 90B   & 13.5 & 71.0 & \textbf{15.6} & -2.1 & - \\
                  & & Gemini 2.5 Flash  & 37.9 & \textit{36.8} & 25.3 & 12.6 & -  \\
                  & & GPT-5 Mini   & \textit{39.3} & 43.9 & 16.8 & \textit{22.5} & - \\
\cmidrule{2-8}
& \textbf{Task 2}   
                  & Llama 3.2 90B    & 30.1 & 50.1 & 19.8 & 10.3 & - \\
                  & & Gemini 2.5 Flash  & \textbf{49.9} & \textit{22.6} & 27.5 & 22.4 & -  \\
                  & & GPT-5 Mini   & 48.7 & 34.1 & \textit{17.2} & \textbf{31.5} & -  \\
\midrule
\textbf{Multi-turn} & \textbf{MM-LLM}   
                  & Llama 3.2 90B   & 42.2 & \textit{25.0} & 32.8 & 12.7 & 64.7 \\
                  & & Gemini 2.5 Flash  & 29.2 & 57.5 & \textbf{13.4} & 16.5 & 88.1 \\
                  & & GPT-5 Mini   & \textit{48.9} & 34.0 & 17.1 & \textit{30.4} & \textit{60.8} \\
\cmidrule{2-8}
& \textbf{Task 3}   
                  & Llama 3.2 90B   & 37.1 & 46.1 & 16.9 & 18.9 & 81.7\\
                  & & Gemini 2.5 Flash  & 54.4 & 24.2 & 21.4 & 31.4 & 55.8 \\
                  & & GPT-5 Mini   & \textbf{61.0} & \textbf{22.5} & \textit{16.5} & \textbf{42.5} & \textbf{43.5}\\
\bottomrule
\end{tabular}
\label{tab:baseline_benchmarking_full}
\end{table*}


Tab.\ref{tab:baseline_benchmarking_full} shows the overall performance of MM-LLM-only and straightforward MM-RAG solutions. 
We observe that the best performing MM-LLM-only solution (GPT-5 Mini) achieves only 37\% accuracy and 18\% truthfulness on single-turn QA, and 49\% accuracy and 30\% truthfulness on multi-turn QA, showing CRAG-MM is challenging without RAG. 

For single-turn QA, adding image KG search (Task 1) in a vanilla manner yields only marginal improvement (+4\%). 
Adding web search (Task 2) increases answer accuracy but also slightly increases hallucinations for all models; 
the highest accuracy and truthfulness is only 50\% and 32\% respectively. 
We observe similar trend for multi-turn QA, 
highlighting the gap that still exists after applying straightforward solutions. 


\subsection{RQ2: Industry SOTA and Competition Winning Solutions}
\label{sec:rq2_sota_and_competition}
We report results from two groups of  solutions. First, we report winning solutions in the CRAG-MM leaderboard~\footnote{\url{https://www.aicrowd.com/challenges/meta-crag-mm-challenge-2025/leaderboards}}, where participating teams had three months to build their best-performing systems with the search API described in Section~\ref{sec:retrieval_contents}. These systems generally combined multi-task learning with supervised fine-tuning on the \texttt{Llama-3.2-11B-Vision-Instruct} model~\citep{meta2024llama32}, 
For comparison, we also report results of the straightforward solution using the same base model.
Additionally, we report industry state-of-the-art (SOTA) MM-RAG solutions. We selected three systems built upon MM-LLMs and search engines, queried them with CRAG-MM questions, collected the responses, and evaluated the quality (details in Appendix~\ref{appedix:sota_benchmarking}).   

Tab.\ref{tab:sota_benchmarking} shows the performance of these solutions. 
Note that the truthfulness scores between the straightforward, leaderboard winning team, and SOTA solutions are not fully comparable, as industry solutions use larger models and potentially have access to much richer content repository; results from winning teams were reported on the private test set, while the other solutions used the public test set. But we believe the trend still holds. We made several observations from the results.

First, the SOTA systems achieved comparable truthfulness compared to the straightforward solutions (+0\% for single-turn and +2\% for multi-turn), but with a higher accuracy (+13\% for single-turn and +9\% for multi-turn). However, the hallucination rate is still high (31\%-49\% for single-turn and 26\%-35\% for multi-turn), showing a substantial gap in trustworthy visual QA systems. Second, the winning team's solution, though lower quality compared to industry solutions, achieved substantial improvement over straightforward MM-RAG solutions using the same base model (+28\% for single-turn and +18\% for multi-turn). They also have the lowest hallucination rates, primarily because the models are able to more properly abstain when uncertain; however, this also causes lower accuracy and higher missing rate, showing the direction for improvement.  

\subsection{RQ3. Directions for Quality Improvements}
Finally, we show the quality of the winning and industry SOTA solutions on different slices of CRAG-MM in Fig.\ref{fig:sota_slicing}. 
(See Fig.\ref{fig:baseline_slicing} in Appendix\ref{appendix:baseline_slicing} for slicing result for straightforward solutions.)

First, all SOTA systems evaluated show degraded truthfulness for low quality images (dropped by up to 46\%, Fig.\ref{fig:sota_slicing}a), showing the need for more robustness in image understanding. ``Low-light" and ``occluded" are the most difficult categories with best truthfulness scores 20\% and 24\% only. 
Interestingly, leaderboard winning solutions have fairly stable quality across different categories, showing the possibility of robustness improvements.
Second, entity recognition is much harder when relying solely on visual information compared to leveraging text clues (dropped by up to 37\%, Fig.\ref{fig:sota_slicing}b). The systems also face much larger challenges when answering questions regarding entities that are less popular (Fig.\ref{fig:sota_slicing}c), both because of the difficulties in entity recognition, and because of lack of abundant information.
Third, MM-RAG systems often face challenges in answering questions requiring specific information (Simple-knowledge), synthesizing multiple pieces of information (comparison and aggregation), and multi-hop information retrieval (Fig.\ref{fig:sota_slicing}d), showing challenges to build MM-RAG systems and substantial room for improving the retrieval, recognition and reasoning ability of the model. 
Finally, having smooth multi-turn conversation remains to be a grand challenge. Consistent with~\citet{laban2025llms}, almost all SOTA systems find it difficult to answer questions that require context from conversation history (dropped by up to 22\%, Fig.\ref{fig:sota_slicing}e). Even the best performing system still has 27\% conversation sessions being early stopped  (Tab.\ref{tab:sota_benchmarking}), and has on average only 3.2 turns successful among 4.9 in total (Fig.\ref{fig:sota_slicing}f).

\begin{table*}[t]
  \centering
  \captionsetup{justification=centering}
  \caption{Performance of winnings' solutions and industry SOTA MM-RAG systems on CRAG-MM.}
  \small
\begin{tabular}{llrrrrrr}
\toprule
  & \textbf{System}   & \textbf{Acc.}  & \textbf{Miss.} & \textbf{Hallu.}     & \textbf{Truth.}  & \textbf{Early Stop.}   \\
\midrule
   \textbf{Single-turn} & Llama 3.2 11B RAG              & \textit{35.3} & \textit{20.8} & 43.9 & -8.6 & - \\
               & Winning team & 29.3 & 61.2 & \textbf{9.6} & \textit{19.7} & - \\
               \cmidrule{2-7}
               & Claude Sonnet   & 45.7 & 5.8 & 48.5 & -2.8 & - \\ 
               & Gemini          & 58.2 & \textit{3.1} & 38.7 & 19.5 & - \\
               & GPT-5           & \textbf{62.7} & 6.8 & \textit{30.5} & \textbf{32.2} & - \\
\midrule
\textbf{Multi-turn}   & Llama 3.2 11B RAG  & \textit{46.6} & \textit{11.0} 
             &  42.4  & 8.8 & \textit{63.7} \\
             & Winning team & 44.4 & 40.2 & \textbf{15.4} & \textit{26.6} & 65.9 \\
             \cmidrule{2-7}
             & Claude Sonnet     & 58.7 & 6.6 & 34.6  & 28.6 & \textbf{25.0} \\ 
             & Gemini            & 66.2 & \textit{3.5} & 30.3 & 30.1 & 37.0 \\
             & GPT-5             & \textbf{70.0} & 3.9 & \textit{26.1}  & \textbf{45.0} & 26.9 \\
\bottomrule
\end{tabular}
\label{tab:sota_benchmarking}
\end{table*}

\begin{figure}[H]
  \centering
  \includegraphics[width=\columnwidth]{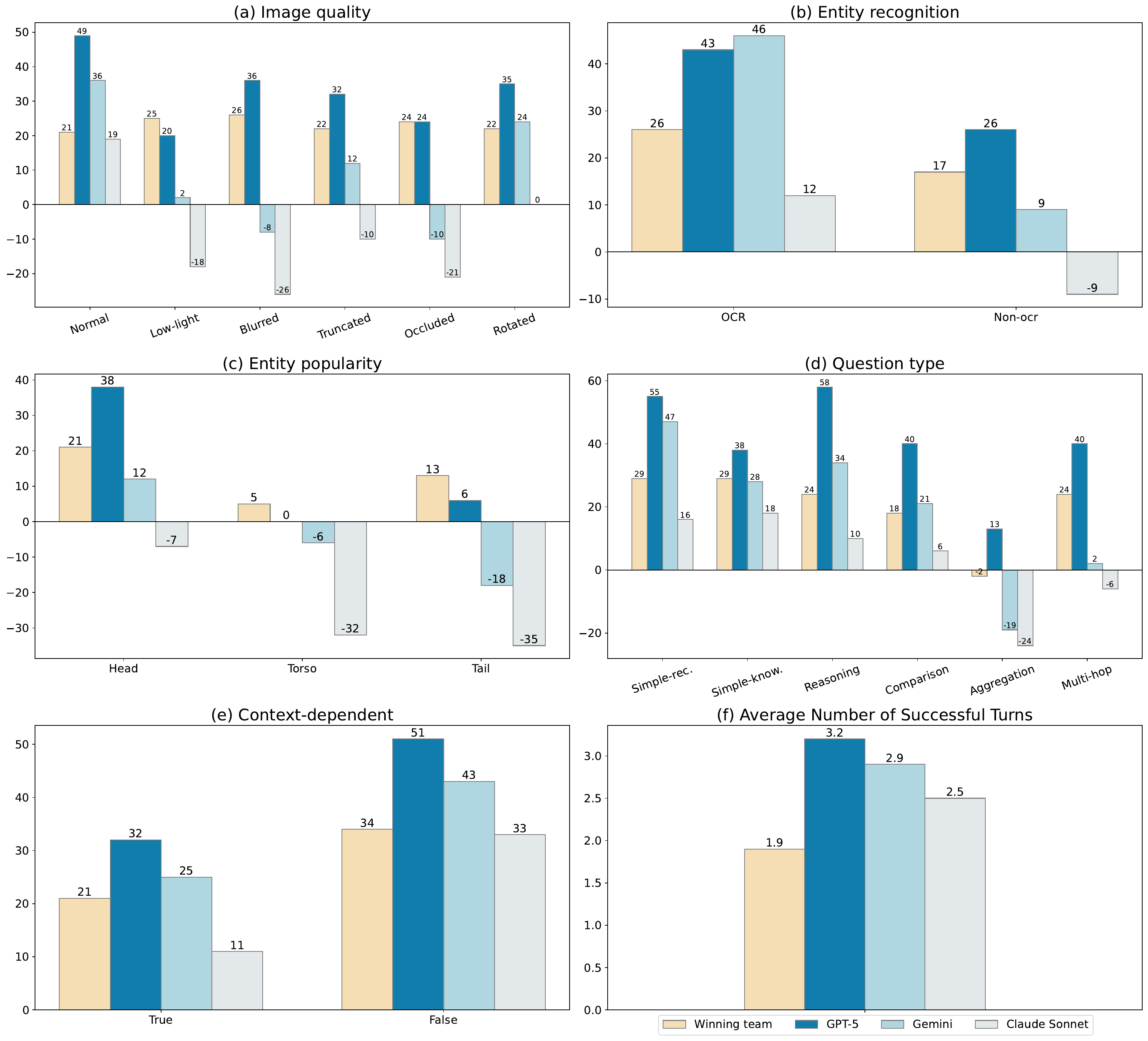}
  \caption{Winning and SOTA systems' performance across different dimensions. Figure (a)--(e) show truthfulness; (f) shows the average number of successful turns in multi-turn QA.}
  \label{fig:sota_slicing}
\end{figure}


\section{Conclusion}
\label{conclusion}
In this paper, we propose CRAG-MM, the first MM-RAG benchmark tailored to wearable AI applications. Its principled design and accessible search APIs empower systematic evaluation of MM-RAG capabilities and inform directions for future advances.

\newpage
\bibliographystyle{assets/plainnat}
\bibliography{paper}

\clearpage
\newpage
\beginappendix

\section{Dataset}
\label{appendix:dataset}

\subsection{Definition of Image Quality}
\label{appendix:image_quality_definition}

\begin{table}[h!]
\centering
\small
\captionsetup{justification=centering}
\caption{Definition of low quality images.}
\label{tab:image_quality_definition}
\begin{tabular}{ll}
\toprule
\textbf{Category} & \textbf{Definition}                                                                                                                        \\
\midrule
Low-light         & \begin{tabular}[c]{@{}l@{}} The entity referred in the query is illuminated by a small amount of light, such as at dusk,\\ during night time, or in shadowed areas. \end{tabular}      \\
\midrule
Blurred     & \begin{tabular}[c]{@{}l@{}} The entity referred in the question appears to be fuzzy, smeared, or indistinct in the image. \end{tabular}         \\
\midrule
Truncated     & \begin{tabular}[c]{@{}l@{}}The entity referred in the query is partially out of the image. \end{tabular} \\
\midrule
Occluded            & \begin{tabular}[c]{@{}l@{}} There is something, like a finger, in the way of the camera and the entity. \\ E.g. the entity is behind a chain link fence or a window screen. \end{tabular}            \\
\midrule
Rotated            & \begin{tabular}[c]{@{}l@{}} The entity referred in the query is more than 10 degrees off from a right orientation. \end{tabular}            \\
\bottomrule
\end{tabular}
\end{table}



\subsection{QA Data Creation}
\label{appendix:qa_data_creation}
\subsection{QA set constructed from KGs}
\label{sec:kg_supported_qa}
Starting from the entities sampled during the image collection, we created a set of simple-knowledge QA data following three steps.
\begin{enumerate}
    \item For each entity type, we selected a meaningful relation $(e, r)$ and created question templates. For example, for \textit{(product, brand)}, we create a question \textit{"what is the brand of this product?"}.
    \item We then used entities sampled for the image collection to pair up with the question. We created equal number of questions for head, torso and tail entities. 
    \item Last, we took the associated attribute values as the answer to the question to obtain question answer pairs.
\end{enumerate}

\smallskip
We created the {\em Comparison, Aggregation, and Reasoning} questions in a similar way but made sure the sampled subject entities can form meaningful questions given the question type. We used heuristics to select entity types for such questions.

Finally, we created multi-hop questions in 3 steps, similar to ~\citet{talmor2018web}. We first sampled an entity $e_1$ from the KG, and selected 2 relation triplets following a two-hop path: $(e_1, r_1, e_2)$ and $(e_2, r_2, e_3)$. We then created a question template describing the path. 
For example, for path \textit{(book$_1$, author, person)} followed by \textit{(author, write, book)}, we created the template \textit{"what is the latest book of this book's author?"}. The answer to the question will be contained in $e_3$ in the second triplet.

\subsection{QA pairs constructed from web contents}
\label{sec:qa_pair_construction_from_web}
We created QA pairs from web search results, in three steps. 

\begin{enumerate}
    \item Given the collected images, we asked annotators to write down questions that are plausible to interact with wearable devices and could possibly be answered by web search. For example, \textit{"how much is this vehicle?"} (where the image is showing a 2024 volkswagon tiguan). 
    \item Next, the annotators will create web search queries based on the image and the question, such as \textit{"how much is the 2024 volkswagon tiguan?"} for the above example. Then they will search the web using the search query to find relevant result for answering the question. Note that we only do text-based web search and rely on image search to provide the entity name. 
    \item Finally, the annotators note down the ground truth answers based on the web search results.
\end{enumerate}

\subsection{Search API Creation}

\subsubsection{Image search API creation}
\label{appendix:image_search_api_creation}
We constructed the image-based mock KG in four steps. 

\begin{enumerate}
    \item We first included the images used to generate the questions in Section~\ref{sec:kg_supported_qa}, and defined coverage categories based on these seed images.
    \item Second, we collected candidate data from the open web (e.g., Wikipedia), including fine-grained entities within each category, their metadata, images, and inter-entity relationships.
    \item From this pool, we selected candidate images to populate the mock KG. To ensure entity coverage, we added positive examples. To create a more realistic testing environment, we also added hard negatives, i.e. images visually similar to the query images but associated with different entities. Formally, let $\mathbf{I}$ denote the set of images for the entity set $\mathbf{E}$ that we used to construct the KG question answer data. For each $i\in \mathbf{I}$, we added up to $30$ hard negatives, denoted by $S(i)$. We further augmented the set with randomly sampled images for the entities in $S(i)$, denoted as $E(S(i))$, i.e., $\mathbf{E} \cup \bigcup_{i\in \mathbf{I}} E(S(i))$. 
    \item Finally, we removed the query images, simulating a realistic setting in which user queries are absent from the KG, but visually similar or related images remain.
\end{enumerate}

All images in the mock KG were encoded with \texttt{CLIP ViT-L/14@336px}, and we built the image retrieval index using ChromaDB\citep{chroma24}.





\begin{figure}[H]
  \centering
  \includegraphics[width=0.7\columnwidth]{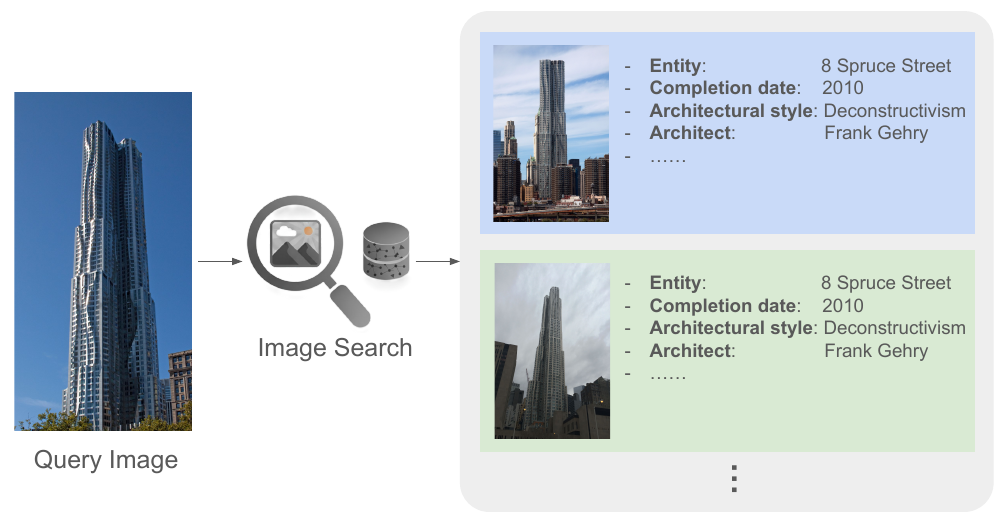}
  \captionsetup{justification=centering}
  \caption{An example of the image search API.}
  \label{fig:image_mock_api_example}
\end{figure}


\subsubsection{Image Search Recall}
\begin{figure}[H]
  \centering
  \includegraphics[width=0.6\columnwidth]{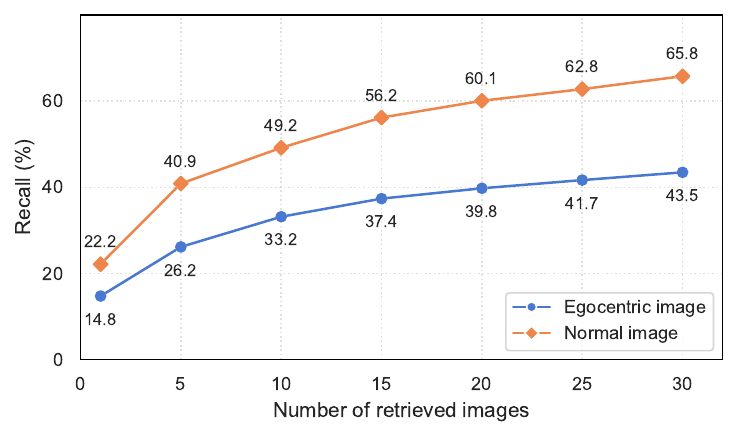}
  \captionsetup{justification=centering}
  \caption{Image search recall by image type using full images to query the index.}
  \label{fig:image_search_recall_by_image_type}
\end{figure}

Fig.\ref{fig:image_search_recall_by_image_type} shows the recall of the image search API for the two image types by using a crude method (query by the original image). Egocentric images has lower recall compared to the normal images, showing the difficulty of entity recognition and image understanding on egocentric images.  

\subsubsection{Web Search API Creation}
\label{appendix:web_search_api_creation}
We created the webpage repository in three steps. 

\begin{enumerate}
    \item We created the ground truth search query and some negative search queries. We constructed the ground truth query by asking the annotators to write down a standalone query for answering the given question regarding the image, without the need to use the image. For example, "who wrote the Martian" for "who wrote this book". To simulate the noise, we also created negative search queries in two ways: 1) a query including the ground truth entity only, 2) queries including similar entities to the ground truth entity. For example, we use Toyota Rav4 if the question is about Volkswagon Tiguan. 
    \item We next used each query to search and store up to $20$ urls via the Brave search API~\cite{brave24}. We collected up to $50$ distinct webpages for each question by combining the search results returned by all the queries and retained the top 50. 
    \item We then pooled the webpages for all questions together and constructed a mock web search API using ChromaDB.
\end{enumerate}

\subsubsection{Unified Search API}
We built a light-weight Python package, providing a unified API to support image and web search. The API can take image URL, PIL image and text query and decide which index (image or web search index) to query depending on the input data type. Below are examples of how to use the search API.
\vspace{\baselineskip}
\begin{lstlisting}
# image search example
results = search_pipeline(image, k = 1)
for result in results:
    print(result)
    print('\n')
# output
{'index': 17030, 'score': 0.906402587890625, 'url': 'https://upload.wikimedia.org/wikipedia/commons/3/34/The_Beekman_tower_from_the_East_River_%286215420371%29.jpg', 'entities': [{'entity_name': '8 Spruce Street', 'entity_attributes': {'name': '8 Spruce Street<br />(New York by Gehry)', 'image': '8 Spruce Street (01030p).jpg', 'image_size': '200px', 'address': '8 Spruce Street<br />[[Manhattan]], New York, U.S. 10038', 'mapframe_wikidata': 'yes', 'coordinates': '{{coord|40|42|39|N|74|00|20|W|region:US-NY_type:landmark|display|=|inline,title}}', 'status': 'Complete', 'start_date': '2006', 'completion_date': '2010', 'opening': 'February 2011', 'building_type': '[[Mixed-use development|Mixed-use]]', 'architectural_style': '[[Deconstructivism]]', 'roof': '{{convert|870|ft|m|0|abbr|=|on}}', 'top_floor': '{{convert|827|ft|abbr|=|on}}', 'floor_count': '76', 'floor_area': '{{convert|1000000|sqft|m2|abbr|=|on}}', 'architect': '[[Frank Gehry]]', 'structural_engineer': '[[WSP Group|WSP Cantor Seinuk]]', 'main_contractor': 'Kreisler Borg Florman', 'developer': '[[Forest City Ratner]]', 'engineer': '[[Jaros, Baum & Bolles]] (MEP)', 'owner': '8 Spruce (NY) Owner LLC', 'management': 'Beam Living', 'website': '{{URL|https://live8spruce.com/}}'}}]}

# web search example (results partially shown)
query='What to know about Andrew Cuomo?'
results = search_pipeline(query, k=1)
for result in results:
    print(result)
    print('\n')
# output
{'index': 'https://en.wikipedia.org/wiki/Mario_Cuomo_chunk_2', 'score': 0.5727531909942627, 'page_name': 'Mario Cuomo - Wikipedia', 'page_snippet': 'He vigorously attacked Ronald Reagan&#x27;s ... brought him to national attention, most memorably saying: &quot;There is despair, Mr. President, in the faces that you don&#x27;t see, in the places that you don&#x27;t visit, in your shining city.&quot; He was immediately considered one of the frontrunners for the Democratic ...He vigorously attacked Ronald Reagan\'s ... brought ......', 'page_url': 'https://en.wikipedia.org/wiki/Mario_Cuomo'}
\end{lstlisting}

\section{Evaluation}
\label{appendix:evaluation_section}

\subsection{Auto-evaluation with LLM-as-a-judge}
We used GPT-4o to evaluate whether the responses are correct, incorrect or missing, given the queries and ground truth answers. To encourage concise answers, we truncate the answers to 75 tokens for auto-evaluation. Tab.\ref{tab:auto-eval_accuracy} shows the overall and per bucket performance. The judge achieves 99\% average accuracy and 91\% average F1 score on the three buckets. 


\label{appendix:evaluation}
\begin{table*}[h]
    \captionsetup{justification=centering}
    \caption{Performance of the GPT-4o based auto-evaluation judge.}
    \label{tab:auto-eval_accuracy}
\centering
\begin{tabular}{lrrrrrr}
\toprule
            &  \textbf{Accuracy} & \textbf{Precision} & \textbf{Recall} & \textbf{F1 score}   \\
\midrule
\textbf{Accurate}   & 98.8 & 98.1 & 90.8 & 94.3 \\
\textbf{Incorrect}   & 98.7 & 68.1 & 91.7 & 78.2 \\
\textbf{Missing}  & 100.0  & 100.0 & 100.0 & 100.0 \\
\midrule
\textbf{Average} & 99.1 & 88.7 & 94.2 & 90.8 \\
\bottomrule
\end{tabular}
\end{table*}

Below is the prompt used for auto-evaluation. We include $\sim$20 in-context examples to improve the judge performance but only show one example due to limited space.

\begin{prompt}[title={Auto-evaluation prompt}]
\begin{lstlisting}
You will be given a question, a ground truth answer, and a model prediction. Your task is to judge if the prediction is correct or not based on the ground truth answer.
## Instructions
Read the question, ground truth answer, and model prediction carefully. Follow the step by step guideline below to make a judgment. 
1. If the prediction indicates uncertainty or refusal to answer, output "Result: WRONG".
2. If the prediction exactly matches the ground truth, output "Result: CORRECT".
3. If the ground truth is a number
    3.1 If the prediction gives a number that almost exactly matches the ground truth, output "Result: CORRECT". 
    3.2 If the prediction gives a number that is not the same as the ground truth, output "Result: WRONG".
4. If the prediction is self-contradictory, output "Result: WRONG".
5. If the prediction is not answering the question, output "Result: WRONG".
6. If ground truth contains a set of objects, 
    6.1 if the prediction contains exactly same objects as the ground truth, output "Result: CORRECT".
    6.2 if the prediction contains different objects from the ground truth, output "Result: WRONG".
    6.3 if the prediction is almost same as the ground truth, use your best judgement to give output.
7. If the prediction is grounded by the ground truth, output "Result: CORRECT".
8. If the prediction is unrelated or contradictory to the ground truth, output "Result: WRONG".
## Additional Guidelines
- Take it as granted that the ground truth is always correct.
- If the prediction gives extra information that is not in the ground truth, it is still correct as long as it is grounded by the ground truth.
- Be careful about numbers. 1 mile is about 1.60934 km. 1 foot is about 0.3048 m. 1 inch is about 2.54 cm. 1 yard is about 0.9144 m. 1 pound is about 0.453592 kg. 1 gallon is about 3.78541 liters. 1 ounce is about 28.3495 grams.

## Output Format
Your judgment should first provide a VERY-SHORT explanation on your rationale. When relevant, you need to include the guidelines above to explain your judgment. Finally, your judgment should clearly state "Result: CORRECT" or "Result: WRONG".

Below are some examples:
EXAMPLES START
Question: who will win the game?
Ground Truth: Lakers is favored to win the game.
Prediction: Sorry, it is hard to predict the outcome of the game.
Explanation: The prediction indicates it is not sure about the answer. So the prediction is incorrect according to the guideline 1.
Result: WRONG
. . .
EXAMPLES END
\end{lstlisting}
\end{prompt}

\subsection{Manual evaluation}
We use manual evaluation to evaluate the auto-evaluation judge performance and winner selection for the competition (see Sec\ref{sec:rq2_sota_and_competition}). Trained human annotators score the answer with perfect (fully correct), acceptable (useful w. minor non-harmful errors), incorrect (wrong or irrelevant) or missing ({\em ``I don't know''}).

\section{Evaluating straightforward MM-RAG solutions}
\label{appendix:baseline_benchmarking}

\subsection{Implementation details}
\label{appedix:baseline_mmrag_implementation}
\textbf{Single-source augmentation (Task 1).} 
We leverage an image search API to retrieve visually similar images and their associated metadata, providing external knowledge to support entity recognition and answer generation, particularly for torso-to-tail entities that MM-LLMs often hallucinate. In our baseline implementation, we use the full input image to query the image search API and retrieve the top 30 images and metadata. A similarity threshold of 0.75 is applied to filter noisy results. The retrieved entity names are retained, and the associated metadata is truncated to 2k tokens. The resulting image search content is appended to the model prompt for response generation.

\textbf{Multi-source augmentation (Task 2 and 3).} 
We incorporate web content in addition to image search, implementing a multi-stage pipeline that integrates image search, query rewriting, web search, and answer generation: Specifically:
\begin{enumerate}
    \item We first retrieve the top 30 entities and associated metadata (as in the single-source setting).
    \item Given the input image, query, and retrieved content, we use Llama-3.2-11B-Vision-Instruct to rewrite the original query into a fully specified textual query independent of the image.
    \item The rewritten query is used to query the web search API, returning up to 50 relevant webpages.
    \item We construct the final prompt by concatenating the image search content (up to 2k tokens), web snippets (up to 8k tokens), and the query.
    \item For multi-turn QA, we additionally include the full conversation history, with previous responses generated by the agent under evaluation.
\end{enumerate}

\subsection{Prompt construction}
\label{appendix:baseline-mmrag-prompt}

We show our prompts used to produce the evaluation results in Tab.~\ref{tab:baseline_benchmarking_full}, Tab.~\ref{tab:baseline_benchmarking_st_full_more} and Tab.~\ref{tab:baseline_benchmarking_mt_full_more}. All the models or APIs are evaluated under the same prompt template.

\begin{prompt}[title={Prompt template for MM-LLM only solution, single-turn QA}]
\begin{lstlisting}
<|system|>
You are a helpful assistant that truthfully answers the user question given an image.
Please follow these guidelines when formulating your response:
1. Your response must be grounded in the image and based on factual information.
2. Keep your response concise and to the point. Strive to answer in one sentence. 
3. If you are uncertain or don't know the answer, respond with "I don't know".
<|user|>
Image: <image>
Query: <query>
\end{lstlisting}
\end{prompt}

\begin{prompt}[title={Prompt template for MM-LLM only solution, multi-turn QA}]
\begin{lstlisting}
<|system|>
You are a helpful assistant that truthfully answers the user question given an image. You are able to engage in multi-turn conversations. Answer the new question based on the given image and information extracted from the previous conversations.
Please follow these guidelines when formulating your response:
1. Your response must be grounded in the image and based on factual information.
2. Build upon previous conversations when responding.
3. Keep your response concise and to the point. Strive to answer in one sentence. 
4. If you are uncertain or don't know the answer, respond with "I don't know".
<|user|>
Image: <image>
Conversation history: <conversation_history>
Query: <query>
\end{lstlisting}
\end{prompt}

\begin{prompt}[title={Prompt template for single-turn augmentation}]
\begin{lstlisting}
<|system|>
You are a helpful assistant that truthfully answers the user question given an image.
Please follow these guidelines when formulating your response:
1. Your response must be grounded in the image and based on factual information.
2. Keep your response concise and to the point. Strive to answer in one sentence. 
3. If you are uncertain or don't know the answer, respond with "I don't know".
<|user|>
Image: <image>
Image search context:
Here is a list of entity names retrieved based on visual similarity to the provided image: 
<list_of_entity_names>
Here are some additional attributes for some of the entities. Only incorporate this information into your answer ONLY IF you are confident the referenced entity is in the provided image.
Entity <i>: <entity_name>
Entity attributes: <meta_data>
. . .
Entity <n>: <entity_name>
Entity attributes: <meta_data>
Incorporate these image entity information into your answer ONLY IF you are confident they refer to the exact same entity. Disregard them otherwise.
Query: <query>
\end{lstlisting}
\end{prompt}

\begin{prompt}[title={Prompt template for multi-source augmentation}]
\begin{lstlisting}
<|system|>
You are a helpful assistant that truthfully answers the user question given an image.
Please follow these guidelines when formulating your response:
1. Your response must be grounded in the image and based on factual information.
2. Keep your response concise and to the point. Strive to answer in one sentence. 
3. If you are uncertain or don't know the answer, respond with "I don't know".
<|user|>
Image: <image>
Image search context:
Here is a list of entity names retrieved based on visual similarity to the provided image: 
<list_of_entity_names>
Here are some additional attributes for some of the entities. Only incorporate this information into your answer ONLY IF you are confident the referenced entity is in the provided image.
Entity <i>: <entity_name>
Entity attributes: <meta_data>
. . .
Entity <n>: <entity_name>
Entity attributes: <meta_data>
Incorporate these image entity information into your answer ONLY IF you are confident they refer to the exact same entity. Disregard them otherwise.
You are given snippets from web page search results based on this question: "{query}". These page snippets may or may not contain relevant or truthful information about the question. Incorporate these information into your answer ONLY IF you are confident they address the question. Disregard them otherwise.

Web search context:
<DOC>
Webpage ID: <i>
Title: <page_name>
Web content snippet: <snippet>
</DOC>
. . .
<DOC>
Webpage ID: <n>
Title: <page_name>
Web content snippet: <snippet>
</DOC>
Incorporate these web search results into your answer ONLY IF you are confident they contain relevant or truthful information about the question. Disregard them otherwise.
Query: <query>
\end{lstlisting}
\end{prompt}

\subsection{Performance of straightforward solutions}
\label{appendix:baseline_eval_full}
In addition to the models discussed in Section~\ref{sec:baseline_benchmarking}, we evaluated our straightforward MM-RAG solutions on CRAG-MM with the following MM-LLMs to provide a more comprehensive performance comparison: \texttt{Llama-4-Maverick}~\citep{meta2025llama4}, \texttt{Pixtral-12B-2049}~\citep{agrawal2024pixtral}, \texttt{Qwen-2.5-VL-72B}~\citep{bai2025qwen2}, \texttt{InternVL3.5-38B}~\citep{wang2025internvl3} and \texttt{Claude 3.7 Sonnet}~\citep{anthropic2025claude3_7_sonnet}. Tab.~\ref{tab:baseline_benchmarking_st_full_more} and Tab.~\ref{tab:baseline_benchmarking_mt_full_more} present the evaluation results for single-turn and multi-turn QAs respectively.

\begin{table*}[h]
  \captionsetup{justification=centering}
  \caption{Performance of straightforward MM-RAG solutions on CRAG-MM single-turn QA.}
  \label{tab:baseline_benchmarking_st_full_more}
\centering
\begin{tabular}{llrrrrr}
\toprule
\textbf{Single-turn}  & \textbf{Model} & \textbf{Acc.} & \textbf{Miss.} & \textbf{Hallu.} & \textbf{Truth.}   \\
\midrule
\textbf{MM-LLM}   & Llama 3.2 11B  & 24.4 & 34.4 & 41.3 & -16.9 \\
                  & Llama 3.2 90B  & 28.2 & 39.1 & 32.8 & -4.6 \\
                  & Llama 4 Maverick (FP8) & 33.3 & \textit{11.0} & 55.7 & -22.4 \\
                  & Pixtral 12B 2409           & 21.9 & 36.1 & 42.1 & -20.1    \\
                  & Qwen2.5 VL 72B      & 26.0 & 45.9 & 28.1 & -2.1  \\
                  & InternVL3.5 38B   & 18.0 & 54.7 & 26.6 & -7.9  \\   
                  & Claude 3.7 Sonnet  & 30.9 & 28.9 & 40.2 & -9.3  \\
                  & Gemini 2.5 Flash  & 36.6 & 38.0 & 25.4 & 11.2   \\
                  & GPT-5 Mini   & \textit{37.4} & 43.7 & \textit{19.0} & \textit{18.4}
                  \\
\midrule
\textbf{Task 1} & Llama 3.2 11B  & 19.0 & 52.4 & 28.7 & -9.7 \\
                  & Llama 3.2 90B   & 13.5 & 71.0 & 15.6 & -2.1 \\
                  & Llama 4 Maverick (FP8) & 23.5 & 52.5 & 24.0 & -0.5 \\
                  & Pixtral 12B 2409          & 23.1 & 35.2 & 41.6 & -18.5 \\
                  & Qwen2.5 VL 72B      & 20.3 & 63.1 & 16.6 & 3.7 \\
                  & InternVL3.5 38B  & 13.4 & 74.2 & \textbf{12.4} & 1.0 \\
                  & Claude 3.7 Sonnet  & 24.5 & \textit{21.5} & 54.0 & -29.5 \\
                  & Gemini 2.5 Flash  & 37.9 & 36.8 & 25.3 & 12.6 \\
                  & GPT-5 Mini   & \textit{39.3} & 43.9 & 16.8 & \textit{22.5}              
                  \\
\midrule
\textbf{Task 2} & Llama 3.2 11B   & 35.3 & 20.8 & 43.9 & -8.6 \\
                  & Llama 3.2 90B    & 30.1 & 50.1 & 19.8 & 10.3 \\
                  & Llama 4 Maverick (FP8) & 44.1 & 15.0 & 40.9 & 3.2 \\
                  & Pixtral 12B 2409          & 42.1 & 4.9 & 53.1 & -11.0 \\
                  & Qwen2.5 VL 72B   & 31.8 & 44.7 & 23.6 & 8.2 \\
                  & InternVL3.5 38B  & 37.9 & 34.3 & 27.8 & 10.1 \\
                  & Claude 3.7 Sonnet  & 30.7 & \textbf{0.9} & 68.4 & -37.7 \\
                  & Gemini 2.5 Flash  & \textbf{49.9} & 22.6 & 27.5 & 22.4 \\
                  & GPT-5 Mini   & 48.7 & 34.1 & \textit{17.2} & \textbf{31.5}
                  \\
\bottomrule
\end{tabular}
\end{table*}

\begin{table*}[h]
  \captionsetup{justification=centering}
  \caption{Performance of straightforward MM-RAG solutions on CRAG-MM multi-turn QA.}
  \label{tab:baseline_benchmarking_mt_full_more}
\centering
\begin{tabular}{llrrrrrr}
\toprule
\textbf{Multi-turn} & \textbf{Model} & \textbf{Acc.} & \textbf{Miss.} & \textbf{Hallu.} & \textbf{Truth.} & \textbf{Early Stop.}  \\
\midrule
\textbf{MM-LLM}   & Llama 3.2 11B   & 36.5 & 19.5 & 44.0 & 1.6 & 74.9 \\
                  & Llama 3.2 90B   & 42.2 & 25.0 & 32.8 & 12.7 & 64.7 \\
                  & Llama 4 Maverick (FP8) & 43.4 & 16.0 & 40.6 & 9.8 & 65.7 \\
                  & Pixtral 12B 2409          & 27.9 & 44.2 & 27.9 & 2.5 & 82.3 \\
                  & Qwen2.5 VL 72B      & 40.1 & 25.6 & 34.3 & 12.1 & 71.3 \\
                  & InternVL3.5 38B  & 37.6 & \textit{9.4} & 53.0 & -1.1 & 73.0 \\
                  & Claude 3.7 Sonnet  & 38.2 & 30.2 & 31.6 & 10.9 & 74.6 \\            
                  & Gemini 2.5 Flash  & 29.2 & 57.5 & \textbf{13.4} & 16.5 & 88.1 \\
                  & GPT-5 Mini   & \textit{48.9} & 34.0 & 17.1 & \textit{30.4} & \textit{60.8} 
                  \\
\midrule
\textbf{Task 3} & Llama 3.2 11B   & 46.6 & 11.0 & 42.4 & 8.8 & 63.7 \\
                  & Llama 3.2 90B   & 37.1 & 46.1 & 16.9 & 18.9 & 81.7 \\
                  & Llama 4 Maverick (FP8) & 53.9 & 9.5 & 36.6 & 19.7 & 55.5 \\
                  & Pixtral 12B 2409          & 48.7 & 6.5 & 44.7 & 9.8 & 62.3 \\
                  & Qwen2.5 VL 72B      & 42.0 & 30.2 & 27.9 & 17.0 & 71.0 \\
                  & InternVL3.5 38B  & 50.3 & 16.9 & 32.8 & 20.5 & 58.0 \\
                  & Claude 3.7 Sonnet  & 55.4 & \textbf{5.0} & 39.6 & 19.6 & 49.7 \\
                  & Gemini 2.5 Flash  & 54.4 & 24.2 & 21.4 & 31.4 & 55.8 \\
                  & GPT-5 Mini   & \textbf{61.0} & 22.5 & \textit{16.5} & \textbf{42.5} & \textbf{43.5} 
                  \\
\bottomrule
\end{tabular}
\end{table*}

\subsection{Performance of straightforward solutions on CRAG-MM slices}
\label{appendix:baseline_slicing}

\begin{figure}[h!]
  \centering
  \includegraphics[width=\linewidth]{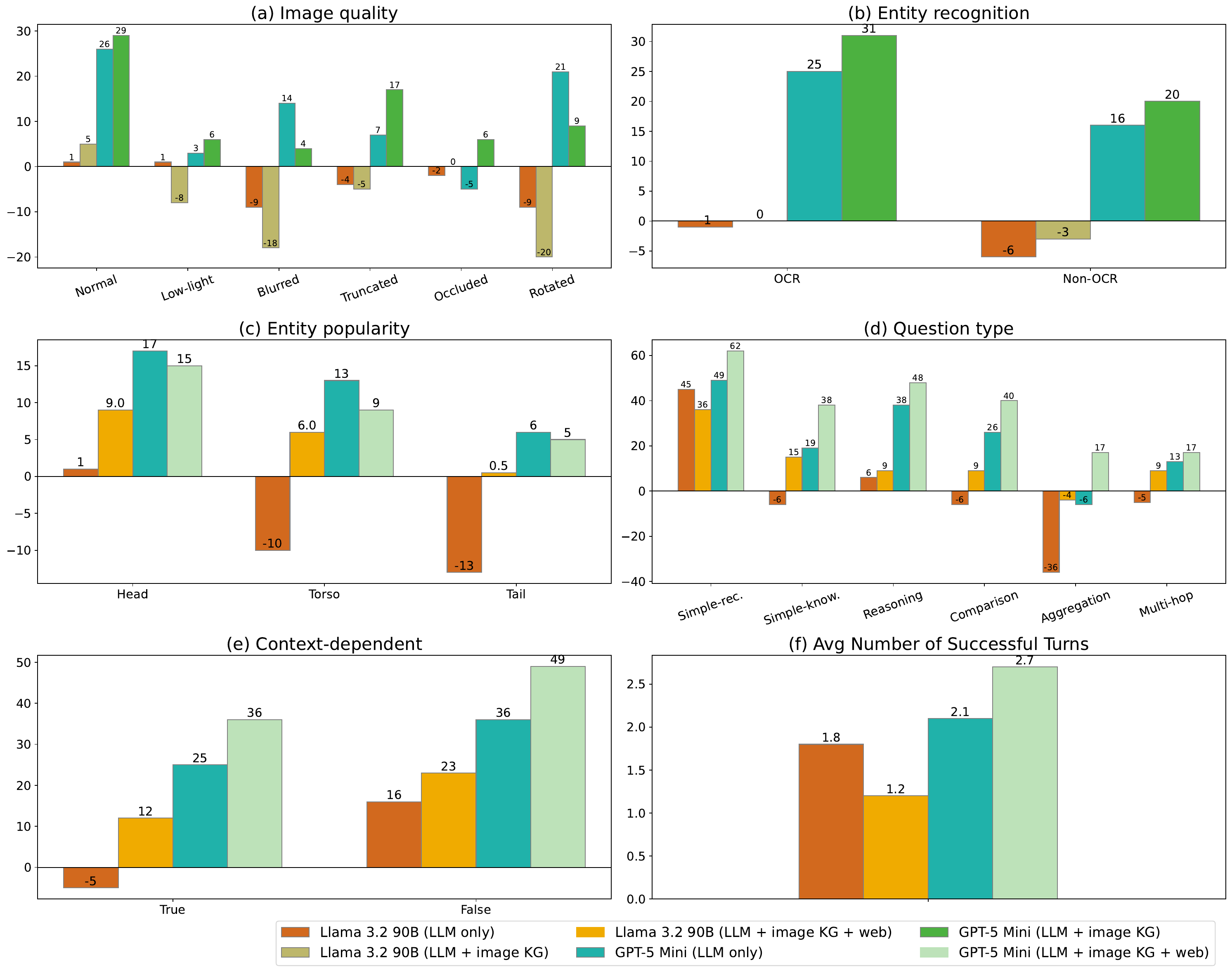}
  \captionsetup{justification=centering}
  \caption{Performance of straightforward MM-RAG solutions across different slices of CRAG-MM. Figure (a)--(e) show truthfulness in percentage; figure (f) shows the average number of successful turns in multi-turn QA.}
  \label{fig:baseline_slicing}
\end{figure}

Fig.\ref{fig:baseline_slicing} shows the performance of the MM-LLM-only and straightforward solutions on different slices of CRAG-MM, showing that CRAG-MM reveals interesting insights and allows large room for improvement for developing MM-RAG systems. 

First, understanding the entity and context in the image is important and challenging for wearable QA, particularly when there is quality issue in the image (Fig.\ref{fig:baseline_slicing}a) or when text is not present - the model needs to recognize the entity purely from the visual information (Fig.\ref{fig:baseline_slicing}b). Adding image search in a crude manner shows mixed effect on different low quality categories (-11\% - +10\%), showing the need for proper pre-processing or smarter image search. 

Second, properly leveraging the retrieved information is non-trivial, particularly when the query involves entities that are less popular (Fig.\ref{fig:baseline_slicing}c), requires external knowledge, or demands synthesizing multiple pieces of information to produce an answer (Fig.\ref{fig:baseline_slicing}d).

Third, results on multi-turn conversation reveals a large gap in conducting smooth conversation - more than 44\% conversations were early stopped (has two consecutive failures, Tab.\ref{tab:baseline_benchmarking_mt_full_more}); the best truthfulness score is only 36\% for questions that require conversation history to answer (Fig.\ref{fig:baseline_slicing}e); on average the best system only has 2.7 successful turns among the 4.9 total for each multi-turn conversation (Fig.\ref{fig:baseline_slicing}f). 

\clearpage

\section{Evaluating industry SOTA solutions}
\label{appedix:sota_benchmarking}

We tested the SOTA MM-RAG systems by sending each question in the public test set as input and collecting the responses for evaluation. Since these SOTA systems all have embedded retrieval systems, the provided retrieval results in CRAG-MM were \textbf{not} used. 

We evaluated three systems: GPT-5-2025-08-07, Gemini-2.5-Pro and Claude-Sonnet-4-20250514, and assessed via the API platforms provided by OpenAI, Google AI and Anthropic, respectively. Tool usage was explicitly enabled to ensure search functionality was available for our end-to-end assessment of MM-RAG systems. Each system was called during the following dates in Pacific Time: 09/01/2025$\sim$09/09/2025 (GPT-5), 09/01/2025$\sim$09/10/2025 (Gemini), and 09/02/2025$\sim$09/10/2025 (Claude Sonnet). 


Below is our parameter setting for tool usage specifically.

\begin{lstlisting}
gpt_tool_calling = [
    {
        "type": "web_search_preview"
    }
]
gemini_generate_content_config = types.GenerateContentConfig(
    thinking_config=types.ThinkingConfig(
        thinking_budget=-1,
    ),
    tools= [
        types.Tool(googleSearch=types.GoogleSearch()),
    ],
)
claude_tool_calling = [
        {
            "name": "web_search",
            "type": "web_search_20250305"
        }
]
\end{lstlisting}


We present the prompts used to produce the SOTA MM-RAG evaluation results in Tab.\ref{tab:sota_benchmarking} and Fig.\ref{fig:sota_slicing} as follows.

\begin{prompt}[title={Prompt template for single-turn QA}]
\begin{lstlisting}
You are a helpful assistant that truthfully answers user questions about the provided image. 
Keep your response concise and to the point. Strive to answer in one sentence. 
If you are unsure or don't know the answer, respond with "I don't know".
Search the internet and provide a short answer to the question: <query>
Image: <image>
\end{lstlisting}
\end{prompt}

\begin{prompt}[title={Prompt template for multi-turn QA}]
\begin{lstlisting}
Conversation history: <conversation_history>
You are a helpful assistant that truthfully answers user questions about the provided image. 
Keep your response concise and to the point. Strive to answer in one sentence. 
If you are unsure or don't know the answer, respond with "I don't know".
Search the internet and provide a short answer to the question: <query>
Image: <image>
\end{lstlisting}
\end{prompt}


\end{document}